\documentclass[reprint,superscriptaddress,showpacs,amsmath,amssymb,aps,pre]{revtex4-1}
\usepackage{graphicx}
\usepackage[justification=raggedright]{caption} % 左对齐

\usepackage{epstopdf}
\usepackage{amsmath}
\usepackage{diagbox} 
\usepackage{booktabs}
\usepackage{array}         % 用于增强表格功能
\usepackage{multirow}     % 用于跨行单元格
\usepackage{makecell}     % 用于在单元格中换行
\usepackage{xcolor}

\usepackage[ruled,vlined]{algorithm2e}

\usepackage{xurl} %允许在 DOI、URL 等连续字符处自动换行，它比 url 宏包更强大
\usepackage[colorlinks=true, citecolor=blue, urlcolor=blue, linkcolor=blue]{hyperref}
\usepackage{subcaption}
\begin{document}

%\title{Fairness Evolution In Multi-Objective Reinforcement Learning}
%\title{The evolution of fairness: a multi-objective reinforcement learning framework}
\title{Evolution of fairness in multi-objective reinforcement learning framework}
%\title{Evolution of fairness from a multi-objective reinforcement learning perspective}
%\title{Evolutionary dynamics of fairness in a multi-objective reinforcement learning framework}
%\title{Emergence of fairness in multi-objective reinforcement learning games}

\author{Jingyi Zhang}
\affiliation{School of Physics and Information Technology, Shaanxi Normal University, Xi'an 710061, P. R. China}
\author{Xin Ou}
\affiliation{School of Physics and Information Technology, Shaanxi Normal University, Xi'an 710061, P. R. China}
\author{Guozhong Zheng}
\affiliation{School of Physical Science and Technology, Inner Mongolia University, Hohhot 010021, China}
\author{Shengfeng Deng}
\affiliation{School of Physics and Information Technology, Shaanxi Normal University, Xi'an 710061, P. R. China}
\author{Jiqiang Zhang}
\affiliation{School of Physics, Ningxia University, Yinchuan 750021, P. R. China}
\author{Li Chen}
\email[Email address: ]{chenl@snnu.edu.cn}
\affiliation{School of Physics and Information Technology, Shaanxi Normal University, Xi'an 710061, P. R. China}

\date{\today}

\begin{abstract}
Fairness, as a fundamental social norm, continues to pose a longstanding puzzle regarding its emergence. Traditional game-theoretic models largely rely on the assumption of \emph{Homo economicus}, wherein individuals are purely rational and self-interested, acting solely to maximize material payoffs. Such accounts, however, overlook the multidimensional nature of human decision-making, which is often shaped also by other considerations beyond economic incentives. To address this gap, we propose a multi-objective reinforcement learning framework that models the evolution of fairness as a dynamic trade-off between material payoff maximization and fairness-driven moral behavior, regulated by a fairness pressure coefficient. Using simulations of a two-objective Q-learning ultimatum game, we find that increased fairness pressure promotes fair outcomes, as expected. Strikingly, however, under moderate pressure, responder behavior reverses: responders become ``forgiving" by accepting low offers -- a pattern in line with our daily experience. Microscopic analyses reveal that this strategy reversal stems from competition between payoff-maximizing and fairness-oriented preferences. We further extend our framework to an asymmetric setting, where proposers and responders assign different weights to the two objectives. Overall, our work expands the reinforcement learning paradigm from a single-objective to a multi-objective formulation, offering a versatile tool for elucidating a broader range of human social behaviors.
\end{abstract}

\maketitle
%------------------------------------------------------------------------------%
\section{introduction}\label{sec:introduction}

Fairness is a foundational pillar of human civilization and a driving force behind the evolution of social cooperation~\cite{Jusup2022Social}. Across historical epochs -- from resource-sharing practices in primitive societies to the institutional architectures of modern states -- the pursuit of fairness has consistently served as a critical mechanism for reconciling individual and collective interests, mitigating social conflicts, and sustaining long-term stability~\cite{Boyd2009Culture,Piketty2014Capital,Stiglitz2019People,Fehr2003Nature}. Understanding how fairness emerges and persists is therefore essential for comprehending the trajectory of human civilization.

Evolutionary game theory offers a powerful framework for studying the origins and maintenance of fairness~\cite{Zhang2025Brief}. Within this theoretical framework, the ultimatum game (UG) has become a canonical model~\cite{Guth1982Experimental,Guth2014More}. In the UG, two players are asked to divide a fixed sum of money: one player, the proposer, suggests a division, while the other, the responder, decides whether to accept or reject it. Acceptance results in the proposed split; rejection leaves both players with nothing. Under the conventional \emph{Homo economicus} assumption~\cite{Samuelson2005Economics,Simon1957Models}, which portrays individuals as fully rational and self-interested, the proposer should offer the minimal possible amount to maximize personal payoff, and the responder, guided by the principle that ``something is better than nothing", should accept any positive offer.

However, a large body of behavioral experiments has consistently revealed substantial deviations from this theoretical prediction: actual offers typically fall between 
$40\%$ and $50\%$ of the total stake, while offers below $20\%$ are frequently rejected~\cite{Kagel1995Handbook,Bolton1995Anonymity,Thaler1988Anomalies,Wang2024SelfServing}. Such robust behavioral patterns indicate that humans exhibit a strong preference for fairness, even at the expense of personal monetary gain. To resolve this gap, various mechanisms have been proposed to account for the emergence and maintenance of fairness~\cite{Debove2016Models}. In particular, previous work has identified factors -- such as population structure~\cite{Page2000Spatial,Cochard2021Social,Kuperman2008Effect,Debove2016ModelsDuplicate}, role assignment~\cite{Chiang2007Evolution,Deng2021Effects,Yang2023Role}, and noise~\cite{Rand2013Evolution,Gale1995Learning} -- as critical determinants. Moreover, individual-level attributes, including reputation~\cite{Zhang2023Reputation,Zhao2025Warmth}, spite~\cite{Forber2014Evolution}, and empathy~\cite{Page2002Empathy,Szolnoki2012Defense}, have also been shown to significantly shape fair behavior. Notably, the majority of these studies operate under the imitation learning (IL) paradigm~\cite{Nowak1992Evolutionary,Szabo1998Evolutionary}, wherein individuals update their strategies by observing and replicating the higher-payoff actions of their neighbors. This mode of decision-making is inherently ``outward-looking", as it relies on social comparison and behavioral imitation.

In recent years, reinforcement learning (RL) has gained prominence as a fundamentally distinct computational paradigm for modeling human behaviors~\cite{Sutton2018Reinforcement,Zheng2026Brief}. In contrast to IL, RL agents refine their strategies iteratively through trial-and-error interactions with the environment, without directly mimicking others. More importantly, RL prioritizes the maximization of long-term cumulative rewards over immediate payoffs -- a feature that is crucial for success in complex human endeavors~\cite{MartinGuerrero2021Reinforcement,Lee2012Neural,Olsson2020Neural}. This paradigm has been successfully applied to explain the emergence of diverse social behaviors in the absence of exogenous incentives, including trust~\cite{Zheng2024DecodingTrust,Zhu2025QLearning,Hu2026HigherOrder}, cooperation~\cite{Xie2026Reinforcement,Zhang2024Emergence,Zheng2024PublicGoods,Ding2023Emergence}, collective action~\cite{LopezIncera2020Development,Wang2023Modeling}, resource allocation~\cite{Zhang2019Reinforcement,Zheng2025Optimal}, and even biodiversity maintenance~\cite{Jiang2026Decoding}.

As a promising new paradigm for understanding diverse systems, the current RL framework is, however, restricted to a single-objective practice, where individuals' decision-making is often devoted to maximizing the accumulated rewards~\cite{Zheng2026Brief}. This fails to capture the complex incentives behind people's decision-making in real-world scenarios. For example, players in the behavioral experiments become very sensitive when moral concern is evoked, and the results change significantly when the UG is framed differently~\cite{Eriksson2017costly}. In many scenarios, individuals have to make trade-offs among different, and sometimes competing, objectives to reach their decisons~\cite{Hayes2022Practical,Coello2007Evolutionary,HernandezDelOlmo2012Emergent}. Therefore, it's reasonable to ask: \emph{can we extend the single-objective RL paradigm to a multiple-objective RL framework}? \emph{And whether such a framework provides new insights into human behaviors}?

In this work, we borrow the idea from the field of RL~\cite{Miettinen2002Scalarizing,Deb2011MultiObjective,Ezzahra2025MultiObjective,Wang2026MultiObjective}, aiming to develop a multi-objective reinforcement learning (MORL) framework and apply it to the emergence of fairness. Specifically, we apply multi-objective Q-learning to the ultimatum game and incorporate both personal payoff and fairness objectives. This integration is motivated by the observations in reality that we humans also care about others'  welfare and emotions alongside our own earnings -- we humans are \emph{Homo duplex}~\cite{durkheim1893division} rather than \emph{Homo economicus}. With this two-objective framework, we reveal that a higher level of fairness is observed compared to the single-objective baseline. Interestingly, the responder shows a non-monotonic strategy change as the fairness pressure increases. We clarify the mechanism behind this by examining the evolution dynamics and the associated Q-tables.

The rest of the paper is organized as follows: 
In Sec. \ref{sec:model}, we introduce the spatial ultimatum game implemented with a multi-objective Q-learning algorithm.
Sec. \ref{sec:result} reports the evolutionary equilibria and the non-monotonic responder strategy flip.
Sec. \ref{sec:mechanism} elucidates the mechanism analysis. 
Sec. \ref{sec:extension} extends the framework setup to an asymmetric scenario. 
Sec. \ref{sec:conclusion} concludes our study.

%-------------------------------------model-----------------------------------------%
\section{model} \label{sec:model}
Consider a two-player ultimatum game (UG) in which one player acts as the proposer and the other as the responder, with roles being fixed throughout the evolutionary process. The total endowment to be allocated is normalized to 1. Let $p_{i} \in [0,1]$ denote the strategy of the proposer $i$, representing the offer made to the responder. Let $q_{j} \in [0,1]$ denote the strategy of responder $j$, representing the minimum offer that player $j$ is willing to accept. When players $i$ and $j$ engage in a given round of the UG, their respective payoffs are given by:
\begin{equation}\label{eq:r1}
	\pi _{p}=\left\{\begin{matrix}
		1-p_{i}, & p_{i}\ge q_{j},\\
		0, &p_{i}< q_{j},
	\end{matrix}\right. \ \ \ \   
	\pi _{r}=\left\{\begin{matrix}
		p_{i}, &  q_{i}\le p_{j},\\
		0, &q_{i}>p_{j},
	\end{matrix}\right.
\end{equation}
where $\pi _{p}$ and $\pi _{r}$ denote the payoffs of the proposer and the responder, respectively.

In many behavioral experiments, the offer $p$ and acceptance threshold $q$ are typically configured to three discrete levels: $l<0.5$ (low), $m=0.5$ (medium), and $h>0.5$ (high)~\cite{Treiman2024Consequences,Moretti2010Disgust,Wang2024Rejecting,Wang2024SelfServing}. Under the \emph{Homo economicus} assumption, which posits that individuals are fully rational and self-interested, the proposer is motivated to minimize the offer to maximize personal gain, while the responder is theoretically compelled to accept any positive offer. Consequently, the Nash equilibrium of the game predicts convergence to the unfair split ($l, l$).

\begin{table}[tbp]
	\centering
	\fontsize{10}{14}\selectfont
	\caption[Q-table]{\textbf{Q-table structure for the ultimatum game}. In the two-player scenario, there are three actions available for both players $\mathcal{A} = \{a_1, a_2, a_3\}$, and states are defined by the joint actions of the proposer and responder in the previous round $(p,q)$, corresponding to the state set $\mathcal{S} = \{ s_1, s_2, \ldots, s_{9} \}$. This Q-table structure applies to both the payoff Q-table $Q_R$ and the fairness Q-table $Q_F$.}
	\setlength{\heavyrulewidth}{1pt}
	\setlength{\lightrulewidth}{1pt}
	\begin{tabular}{>{\bfseries}c|>{\bfseries}c>{\bfseries}c>{\bfseries}c>{\bfseries}c}
		\toprule
		\diagbox [width=5em,trim=l] {State}{Action} & $a_1 (l) $ & $a_2 (m)$ & $a_3 (h)$\\
		\hline
		$s_{1}=(l, l)$ & $Q(s_{1},a_{1})$ & $Q(s_{1},a_{2})$ & $Q(s_{1},a_{3})$ \\
		$s_{2}=(l, m)$ & $Q(s_{2},a_{1})$ & $Q(s_{2},a_{2})$ & $Q(s_{2},a_{3})$ \\
		$s_{3}=(l, h)$ & $Q(s_{3},a_{1})$ & $Q(s_{3},a_{2})$ & $Q(s_{3},a_{3})$ \\
		$s_{4}=(m, l)$ & $Q(s_{4},a_{1})$ & $Q(s_{4},a_{2})$ & $Q(s_{4},a_{3})$ \\
		$s_{5}=(m, m)$ & $Q(s_{5},a_{1})$ & $Q(s_{5},a_{2})$ & $Q(s_{5},a_{3})$ \\
		$s_{6}=(m, h)$ & $Q(s_{6},a_{1})$ & $Q(s_{6},a_{2})$ & $Q(s_{6},a_{3})$ \\
		$s_{7}=(h, l)$ & $Q(s_{7},a_{1})$ & $Q(s_{7},a_{2})$ & $Q(s_{7},a_{3})$ \\
		$s_{8}=(h, m)$ & $Q(s_{8},a_{1})$ & $Q(s_{8},a_{2})$ & $Q(s_{8},a_{3})$ \\
		$s_{9}=(h, h)$ & $Q(s_{9},a_{1})$ & $Q(s_{9},a_{2})$ & $Q(s_{9},a_{3})$ \\
		\bottomrule
	\end{tabular}
	\vspace{0cm}
	\label{table:1}
\end{table}

To establish a multi-objective reinforcement learning (MORL) framework, we adopt a single-policy approach~\cite{Gabor1998MultiCriteria,Mannor2004Geometric,VanMoffaert2013Scalarized,Miettinen2002Scalarizing}. In this approach, the algorithm maintains an independent Q-value estimate \(Q_o(s, a)\)for each objective \(o\), forming a Q-vector:
 \begin{equation}\label{eq:r2}
	\mathbf{Q}(s, a) = (Q_1(s, a), \ldots, Q_m(s, a)),
\end{equation}
where $m$ is the number of objectives. This is essentially equivalent to maintaining a separate Q-table for each objective, and the collection of these tables constitutes the multi-objective value estimation. To guide decision-making, these vectorized Q-values must be aggregated into a scalar value. The linear scalarization function accomplishes this by computing a weighted sum of the Q-values across all objectives:
%Essentially, this is equivalent to maintaining a separate Q-table for each objective, which collectively constitute the multi-objective value estimation. To guide decision-making, these vectorized Q-values must be aggregated into a scalar value. The linear scalarization function accomplishes this by computing a weighted sum of the Q-values across all objectives:
\begin{equation}\label{eq:r3}
	\widehat{Q}(s,a) = \sum_{o=1}^{m} w_o \cdot Q_o(s, a).
\end{equation}
where \(w_o\) denotes the weight assigned to objective \(o\), satisfying  \(\sum_{o=1}^{m} w_o = 1\). This weighted summation reflects the decision-maker’s relative preferences toward different objectives. Once the scalar estimate \(\widehat{Q}(s,a)\) is obtained, the procedure is reduced to standard single-objective reinforcement learning.

Specifically, our study employs a multi-objective Q-learning algorithm to govern the decision-making of both players. Each player maintains multiple individual Q-tables, the number of which equals the number of objectives. These Q-tables share the same structure, as shown in Table~\ref{table:1}; they are two-dimensional matrices, with rows corresponding to states and columns to actions. The action space is defined as $\mathcal{A} = \{ l, m, h \}$, comprising three discrete actions: for the proposer, these correspond to low, medium, and high offers, respectively; for the responder, they correspond to low, medium, and high acceptance thresholds. The state is jointly determined by the player's action and the opponent's action in the previous round, yielding a total of 9 possible states $\mathcal{S} = \{ s_1, s_2, \ldots, s_{9} \}$. 
 $Q(s, a)$ is the state-action value function, representing the expected utility of taking action $a$ in state $s$. A higher Q-value typically indicates a stronger preference for selecting the corresponding action.

Here we consider two objectives for each player in our study: i) maximizing individual payoff, i.e., the payoff defined in Eq.~(\ref{eq:r1}), corresponding to the Q-table $Q_P$; and ii) pursuing distributive fairness, measured by whether the payoffs of both players are equal, corresponding to the Q-table $Q_F$. If the proposer's and responder's payoffs are identical, an additional fairness reward $\pi_f=r>0$ is conferred; otherwise, the reward $\pi_f=0$.
While the first objective is rooted in economic incentives and has been extensively examined in previous studies, the second objective is relevant to emotional or moral incentives.

Initially, the two players randomly select an action from the action set $\mathcal{A}$. In each subsequent round $t$, both players have an exploration probability $\epsilon$ to independently choose a random action $a_t$; otherwise, they follow the guidance of the aggregate Q-value $\widehat{Q}(s, a)$, which is a linear scalarization of the payoff Q-table $Q_P(s, a)$ and the fairness Q-table $Q_F(s, a)$:
\begin{equation}\label{eq:r4}
	\widehat{Q}(s, a) = (1 - \omega) \cdot Q_P(s, a) + \omega \cdot Q_F(s, a),
\end{equation}
where the weighting parameter $\omega\in [0,1]$ characterizes the player's preference intensity for fairness. The extreme case of $\omega=0$ is reduced to the single-objective scenario~\cite{Zheng2025DecodingFairness}.  A higher value of $\omega$ indicates a greater reliance on the fairness Q-table during decision-making. With the aggregate Q-table, players choose the action $a_t$ with the highest Q-value in their current state $s_t$. After making their decisions, they receive both payoff and fairness rewards, and the new state $s_{t+1}$ is then determined.

Finally, each player updates both Q-tables to incorporate the newly acquired experience. Specifically, the corresponding state-action Q-values in their Q-tables are revised using the following Bellman equation:
\begin{equation}\label{eq:r5}
	\begin{aligned}
		Q_{P,F}(s_t,a_t) &\leftarrow (1 - \alpha)Q_{P,F}(s_t,a_t) \\
		                           &+ \alpha \left( \pi + \gamma \max_{a'} Q_{P,F}({s_{t+1},a'}) \right).
	\end{aligned}
\end{equation}
Here, $\pi$ is the immediate reward: for $Q_P$, it is the payoff reward given by Eq.~(\ref{eq:r1}); for $Q_F$, it is the fairness reward $\pi_f$. The learning rate $\alpha \in (0,1]$ controls the contribution of new experience in updating the Q-value, and the discount factor $\gamma \in [0,1)$ quantifies the influence of the maximum future reward, i.e., $\max_{a'} Q_{P,F}({s_{t+1},a'})$. 
It is worth noting that the aggregate Q-table $\widehat{Q}(s, a)$ used for decision-making is not updated directly; instead, it is recomputed at each round according to Eq.~(\ref{eq:r4}) based on the updated $Q_P$ and $Q_F$. The pseudo-code of our multi-objective Q-learning model is presented in Algorithm~\ref{alg:morl}.

In this study, the three discrete levels for both offer $p$ and acceptance threshold $q$ are set to $(l, m, h)=(0.3,0.5,0.8)$. The learning parameters are fixed at $\varepsilon = 0.01$, $\alpha = 0.1$, $\gamma = 0.9$~\cite{Zheng2025DecodingFairness}, which requires players to appreciate both their historical experience (small $\alpha$) and future reward (large $\gamma$).  The primary goal of this study is to investigate the impact of the multi-objective learning framework on the emergence of fairness, and to elucidate fundamental new dynamics, if any, due to multi-objective incorporation.

\begin{algorithm}[tbp]
	\SetAlgoNlRelativeSize{0} 
	\SetInd{0.5em}{0.5em} 
	\SetAlgoNlRelativeSize{-1} 
	\SetNlSty{text}{}{\hspace{0.5em}} 
	\caption{Multi-objective Q-learning for ultimatum game}
	\KwIn{Fairness weight $\omega$, fairness reward $r$, learning rate $\alpha$, discount factor $\gamma$, exploration rate $\epsilon$}
	\KwOut{Action adoption probabilities $\rho$}
	{\bf Initialization}\;
	\For{$i \in \{Proposer, Responder\}$}{
	Initialize payoff Q-table $Q_P$, fairness Q-table $Q_F$ and action randomly\;
    }
	\For{each independent run}{
		\For{each step}{
			\For{$i \in \{Proposer, Responder\}$}{
				Get state $s_i$ according to Table \ref{table:1}\;
				Compute aggregate Q-table according to Eq.~\eqref{eq:r4} for all $a \in \mathcal{A}$\; 
				\eIf{rand() $< \epsilon$}{
					$a_i \leftarrow$ random action from $\mathcal{A}$\;
				}{
					$a_i \leftarrow \arg\max_a \widehat{Q}(s_i, a)$\;
				}
			}
			{\bf Execute UG}\; 
			Get proposer's offer $p$ and responder's threshold $q$\;
			{\bf Get payoff reward}\; 
			\eIf{$p \ge q$}{
				$\pi_{p} = 1-p,\ \pi_{r} = p$\;
			}{
				$\pi_{p} = \pi_{r} = 0$\;
			}
			{\bf Get fairness reward}\;
			\eIf{$\pi_{p} = \pi_{r}$}{
				$\pi_{f} = r$\ for both;
				}{$\pi_{f} = 0$\ for both;}
			Get new state $s_i'$ according to $p$ and $q$ and Table \ref{table:1}\;
			\For{each player $i$}{
				Update payoff Q-table $Q_P$ and fairness Q-table $Q_F$ according to Eq.~\eqref{eq:r5} 
			}
		}
	}
	Compute $\rho$ from recorded actions\;
	\label{alg:morl}
\end{algorithm}

%------------------------Fig. 1---------------------%
\begin{figure*}[htbp]
\centering
\includegraphics[width=1\linewidth]{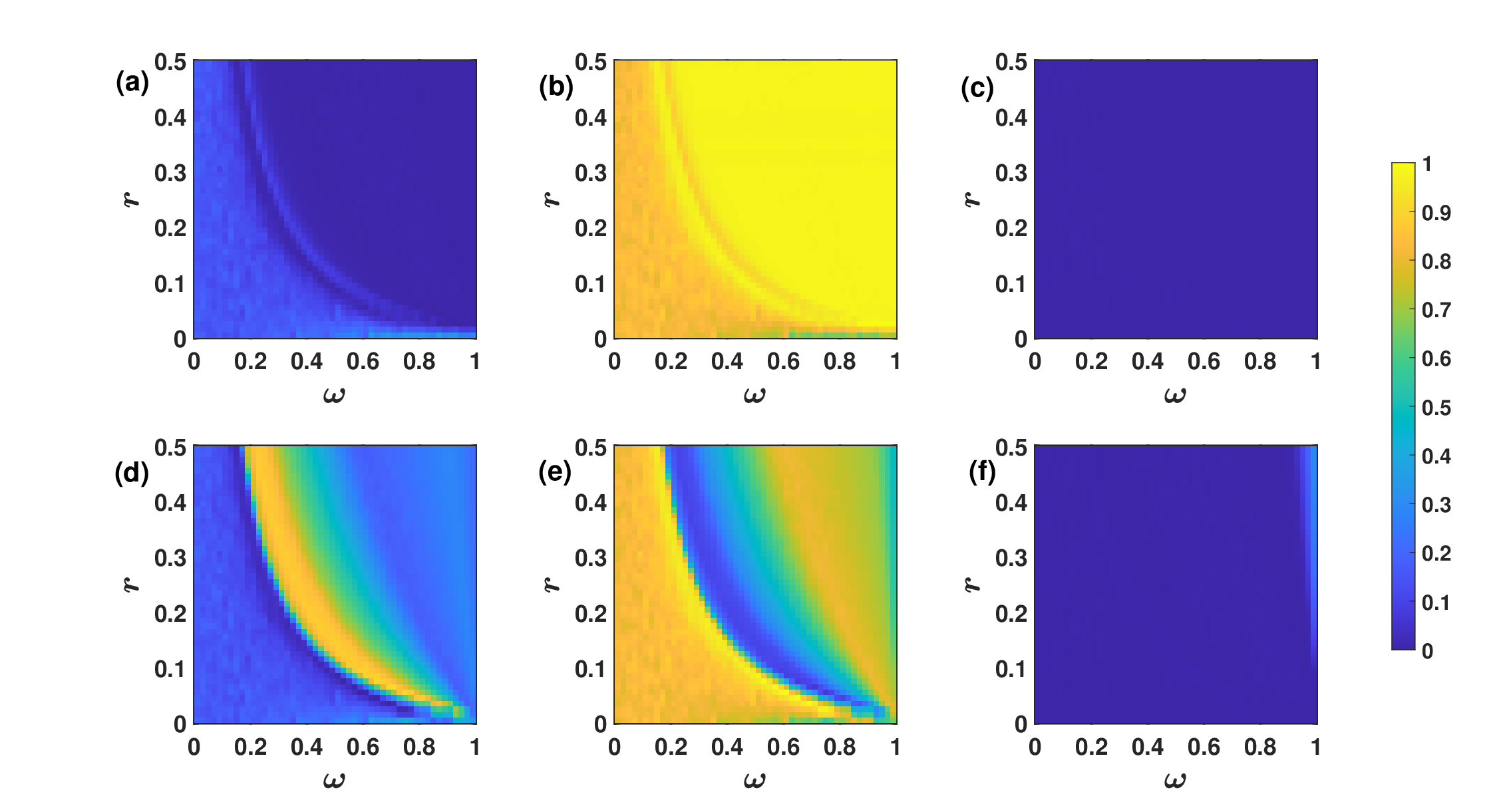}
\caption{\textbf{Phase diagram of probabilities for the three actions with multi-objective Q-learning.}
Probabilities of the action adoption for both proposer and responder are color-coded in the $r-\omega$ parameter space.
Panels (a-c) show the average proportion of the proposer's three actions for $p=l, m, h$, respectively. Panels (d-f) show the average proportion of the responder's three actions, respectively, for $q=l, m, h$. Each data point is obtained from an ensemble average over 100 independent runs. Other parameter: $t_{\text{max}} = 1 \times 10^8$.}
\label{fig:phase_diagram}
\end{figure*}

%--------------------------------------results----------------------------------------%
\section{results}\label{sec:result}

We first report the action adoption probabilities $\rho$ for both the proposer and responder by examining the impact of the two key parameters ($r$ and $\omega$) in the MORL framework, as shown in Fig.~\ref{fig:phase_diagram}. As a benchmark, Ref.~\cite{Zheng2025DecodingFairness} has shown that in the single-objective scenario (i.e., $\omega=1$), the fair strategy $m$ dominates for both proposer and responder, with both adoption probabilities approximately $\rho_{p_m,q_m}\approx85\%$, while the rational strategy ($p_l, q_l$) coexists with the remaining probability.

Figs.~\ref{fig:phase_diagram}(a-c) show that the fair action $p_m$ remains dominant among the three offers, and the adoption probability increases further, with $\rho_{p_m}\rightarrow 1$ as both $r$ and $\omega$ increase. This indicates that the introduction of the fairness objective effectively suppresses the rational strategy, with $\rho_{p_l}\approx 0$. In contrast, Figs.~\ref{fig:phase_diagram}(d-f) reveal that responder behavior exhibits considerable deviation: when both $r$ and $\omega$ are sufficiently low, responders show a marked preference for the acceptance threshold $q_m$; as the two parameters increase, the dominant responder strategy drops to the low acceptance threshold $q_l$; with further increases in $r$ and $\omega$, responders once again favor the intermediate threshold $q_m$. In the extreme case of $\omega\rightarrow1$, players are exclusively fairness-oriented, and proposers consistently offer the fixed amount $p_m=0.5$. Under this condition, any of the three acceptance thresholds $q_{l,m,h}$ yields equal payoffs for both players ($\pi_{p,r}=0.5$ for $q_{l,m}$ and 0 for $q_h$) -- thereby satisfying the fairness criterion -- and thus all three strategies manifest with comparable probabilities. Finally, Figs.~\ref{fig:phase_diagram}(c,f) show that the overgenerous offer is rarely proposed or expected, as the adoption probabilities of both $p_h$ and $ q_h$ remain vanishingly small across the whole domain, except for the extreme case of $\omega\rightarrow1$ for the responder, where even a null reward still yields fairness satisfaction.

%------------------------Fig. 2---------------------%
\begin{figure*}[htbp]
\centering
\includegraphics[width=0.3\linewidth]{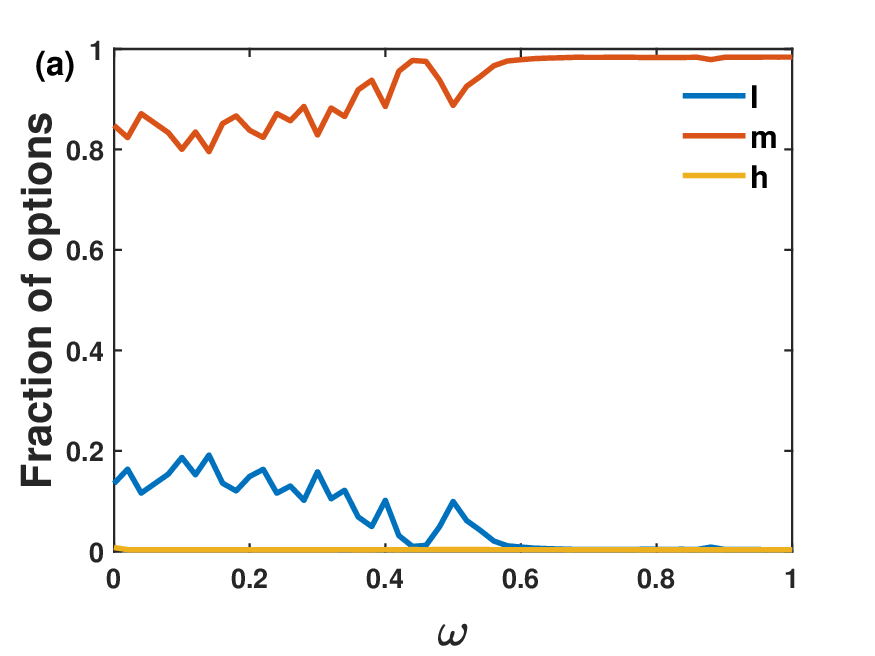}
\includegraphics[width=0.3\linewidth]{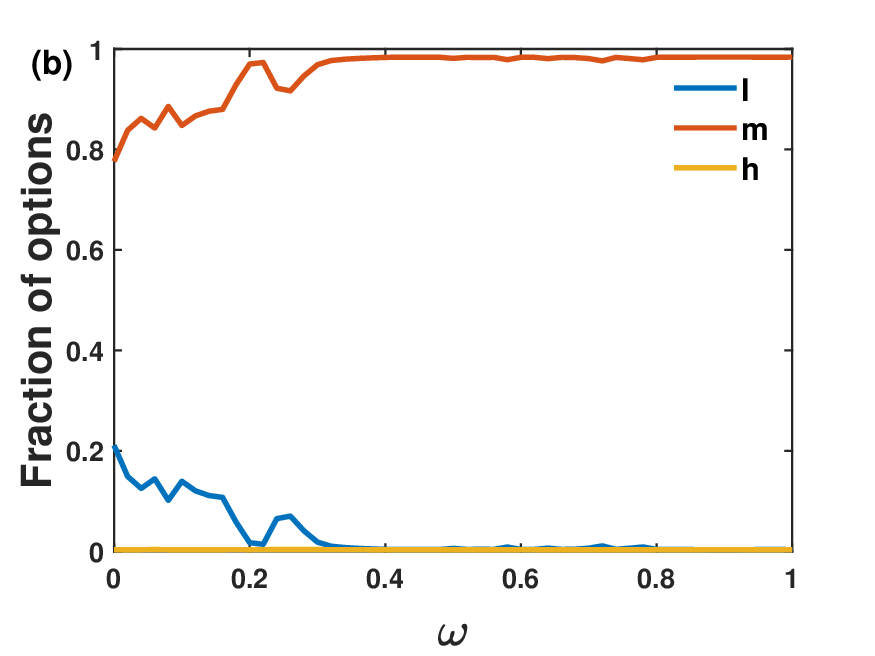}
\includegraphics[width=0.3\linewidth]{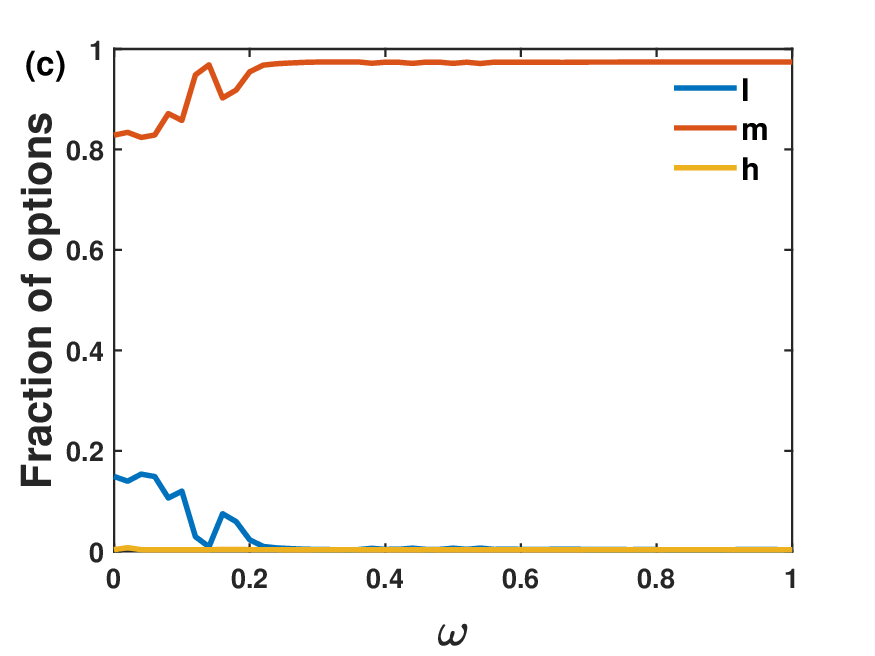}\\
\includegraphics[width=0.3\linewidth]{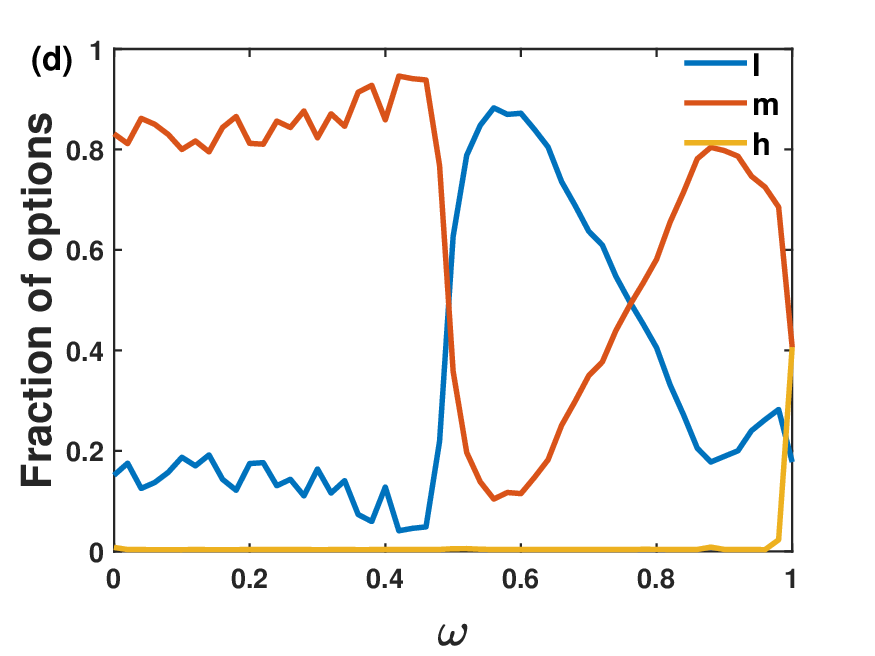}
\includegraphics[width=0.3\linewidth]{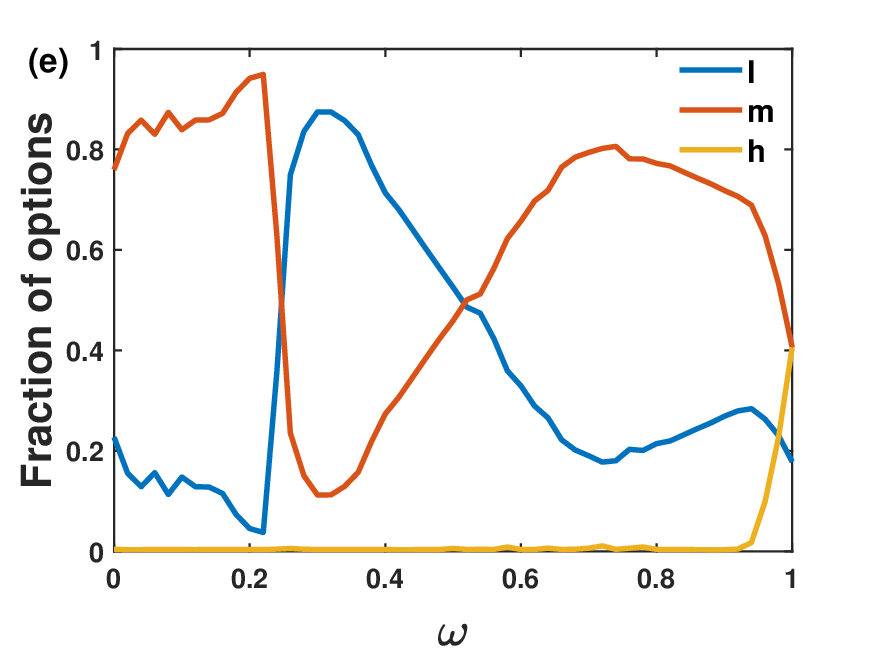}
\includegraphics[width=0.3\linewidth]{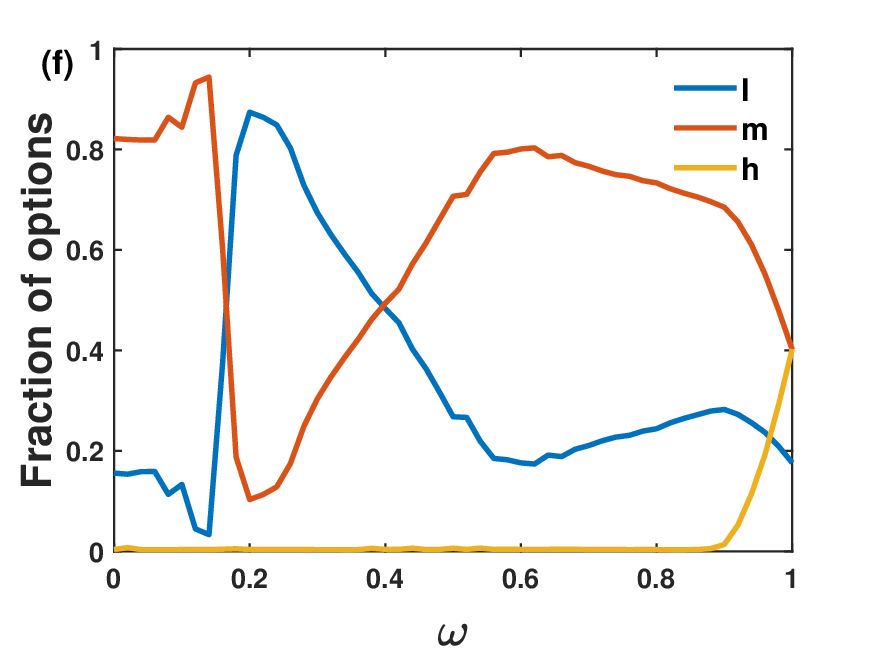}
\caption{\textbf{Dependence of action adoption on the fairness weight $\omega$.}
Panels (a-c) show the proposer's action adoption probabilities for fairness rewards $r=0.1$, 0.3, and 0.5, respectively.
Panels (d-f) are the corresponding results for the responder.
Other parameter: $t_{\text{max}} = 1 \times 10^8$.}
\label{fig:impact_omega}
\end{figure*}
%-----------------------------------------------------------%

To systematically elucidate the non-monotonic dependence of action adoption on the two parameters, we examine the impact of one parameter in detail while fixing the other. Fig.~\ref{fig:impact_omega} shows how the proportions of each action change with the fairness weight $\omega$ for the fairness reward fixed at $r=0.1$, 0.3, and 0.5. The results indicate that increasing the weight of the fairness objective indeed further enhances the adoption of $p_m$, raising it from $\rho_{p_m}\approx85\%$ to nearly full fairness. However, the dependence of the three acceptance thresholds $q_{l,m,h}$ is considerably more complicated. As $\omega$ increases from low values, $q_m$ initially rises like $p_m$; this trend reverses when $\omega$ increases further, with the fraction of $q_m$ dropping and $q_l$ suddenly surging. As $\omega$ continues to increase, the fraction of $q_m$ rises again, and when $\omega\rightarrow 1$, the fractions of the three actions become comparable. Observations across the three columns in Fig.~\ref{fig:impact_omega} show that the dependence curves are qualitatively similar for the three fairness rewards; a larger $r$ shifts the onset of transitions to lower $\omega$ values.

\begin{figure*}[htbp]
\centering
\includegraphics[width=0.3\linewidth]{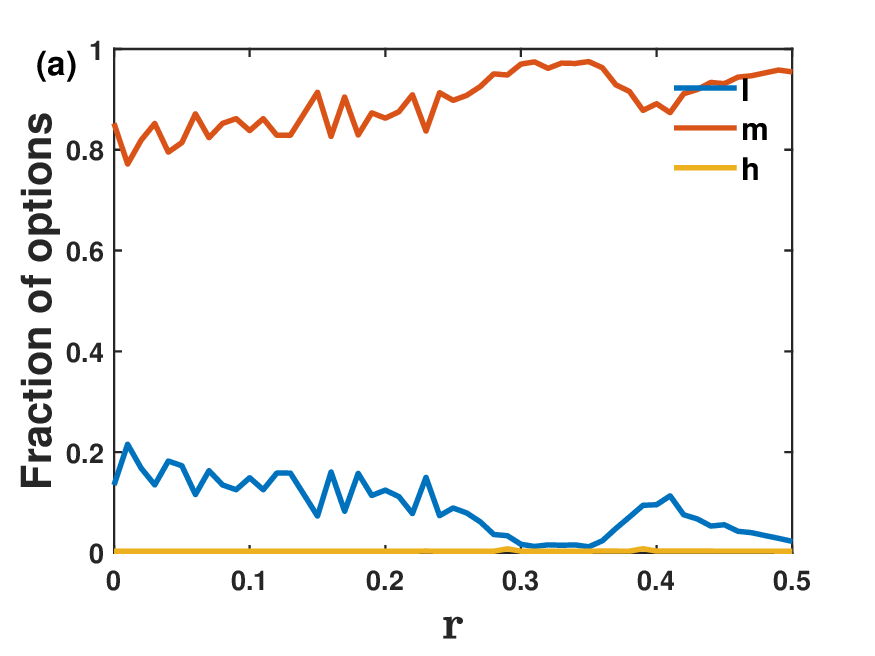}
\includegraphics[width=0.3\linewidth]{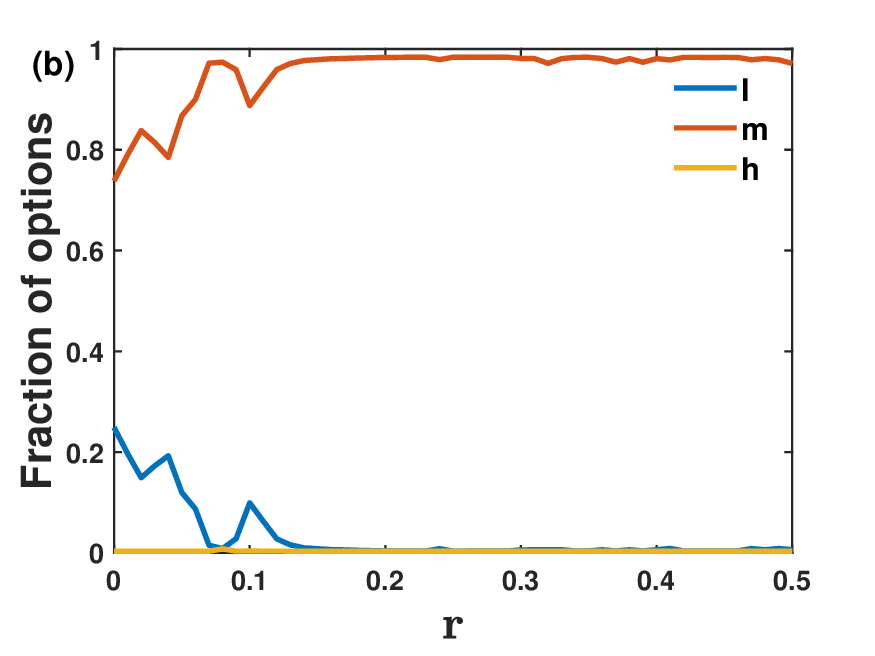}
\includegraphics[width=0.3\linewidth]{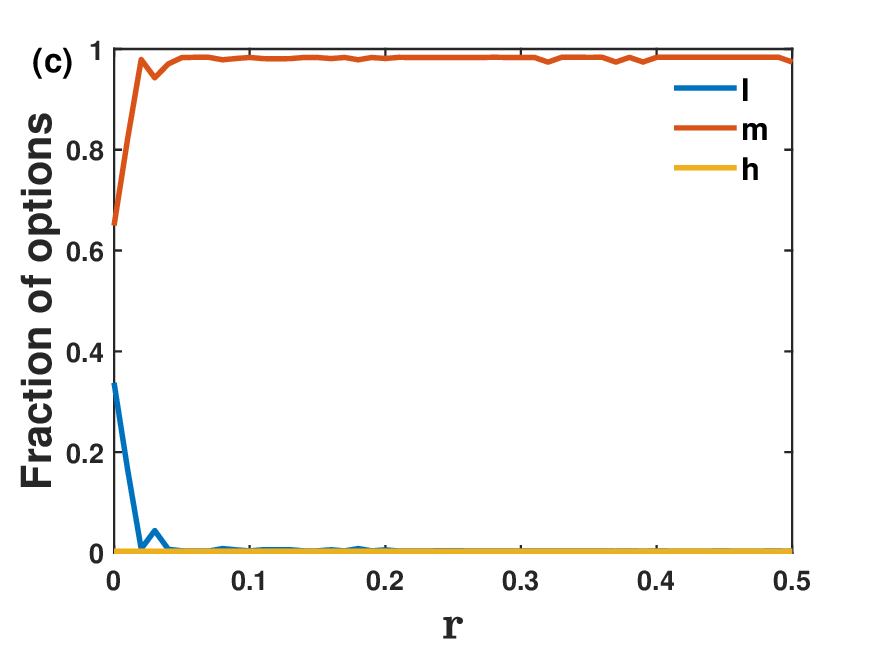}\\
\includegraphics[width=0.3\linewidth]{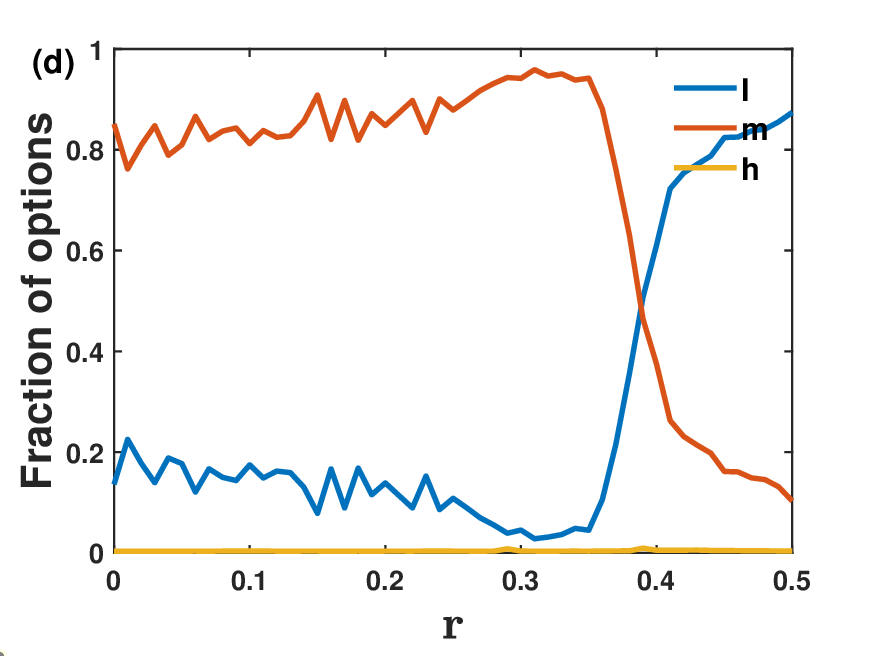}
\includegraphics[width=0.3\linewidth]{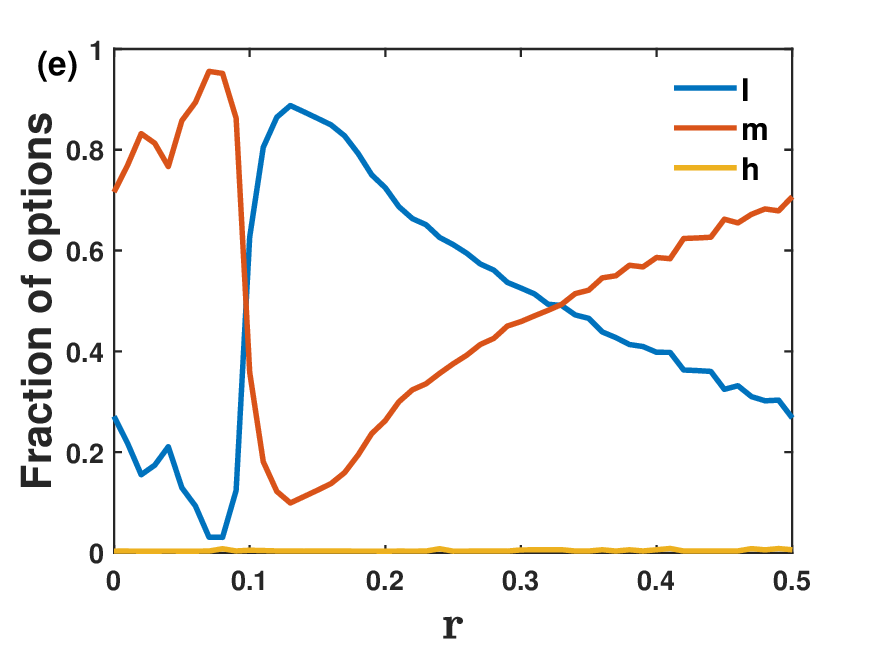}
\includegraphics[width=0.3\linewidth]{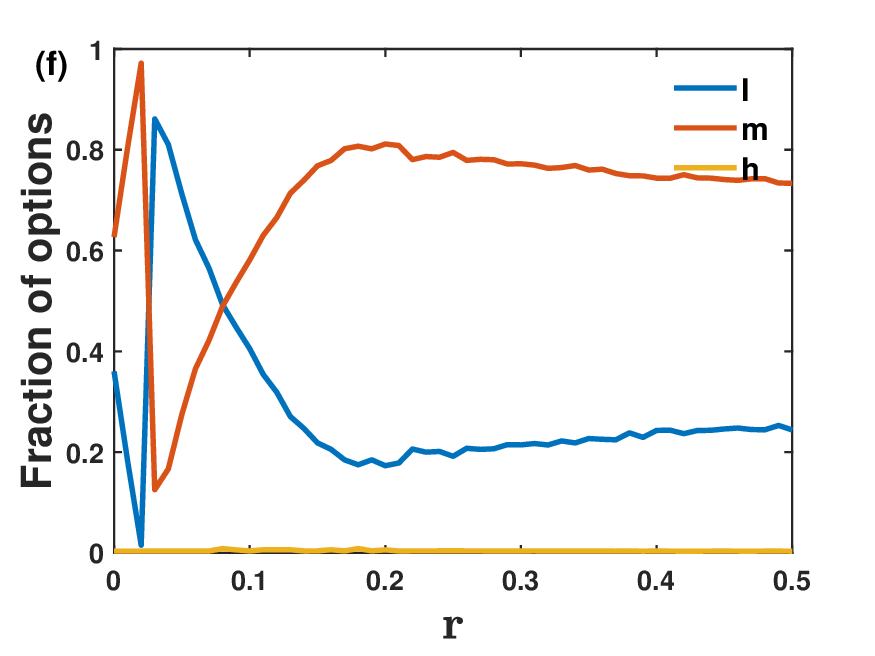}
\caption{\textbf{Dependence of action adoption on the fairness reward $r$.}
Panels (a-c) show the proposer's action adoption probabilities for fairness rewards $\omega=0.2$, 0.5, and 0.8, respectively.
Panels (d-f) are the corresponding results for the responder.
Other parameter: $t_{\text{max}} = 1 \times 10^8$.}
\label{fig:impact_r}
\end{figure*}

This is reasonable as the role of fairness reward $r$ is expected to be qualitatively equivalent to that of the fairness weight $\omega$; a larger value of either $r$ or $\omega$ enhances the relative importance of the fairness objective. Therefore, a similar dependence in the parameter $r$ is anticipated. Fig.~\ref{fig:impact_r} confirms that this is indeed the case, where we fix $\omega=0.2$, 0.5, and 0.8. Figs.~\ref{fig:impact_r}(a-c) show that the fraction of $p_m$ is further promoted as $r$ increases, eventually reaching $p_m\approx1$. Figs.~\ref{fig:impact_r}(d-f) illustrate that the reversal pattern is also present; however, the rising trend of $p_m$ has not yet occurred within the given range of $r$ when $\omega=0.2$, and, conversely, the reversal becomes narrower as $\omega$ becomes too large.  In the subsequent analysis, we focus primarily on how variations in $\omega$ shape the system's evolutionary dynamics, as the fairness reward $r$ yields qualitatively the same impact.

%------------------------Fig. 3---------------------%

%------------------------Fig. 4---------------------%
\begin{figure*}[htbp]
\centering
\includegraphics[width=1.0\linewidth]{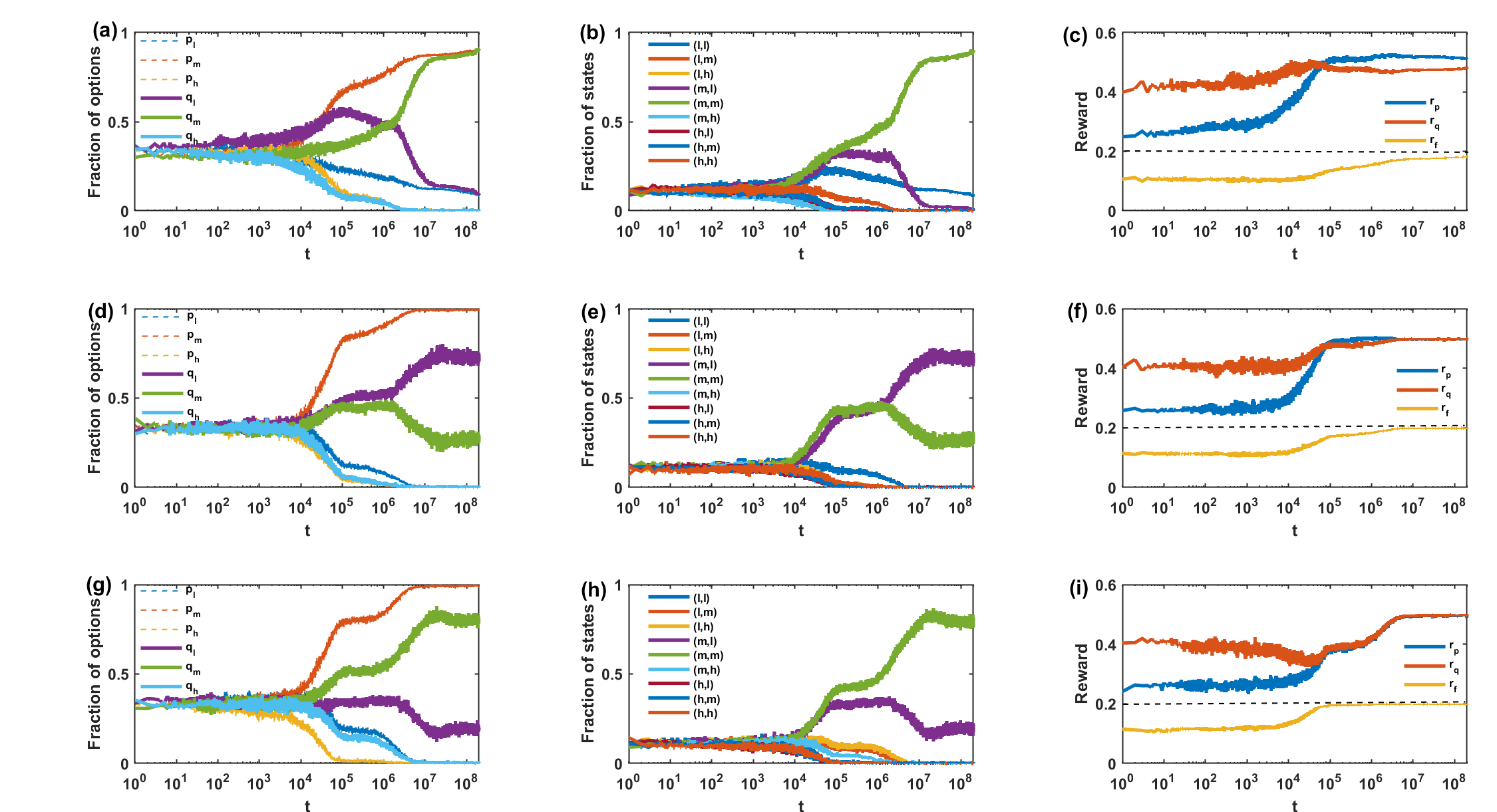}
\caption{\textbf{Evolution at three different fairness pressures.}
The top (a-c), middle (d-f), and bottom (g-i) rows correspond to weak ($\omega=0.2$), intermediate ($\omega = 0.5$), and strong ($\omega = 0.8$) pressure, respectively.
The first column shows time series of the three action fractions for both proposer $p_{l,m,h}$ and responder $q_{l,m,h}$. 
The second column provides time series of all nine state fractions. 
The third column shows time series of proposer and responder payoffs, and the dashed line in each panel corresponds to $r=0.2$, the expected value in a fully fair scenario. 
All data represent ensemble averages over 500 independent runs. 
Other parameter: $t_{\mathrm{max}} = 2 \times 10^8$.}
\label{fig:ts}
\end{figure*}

%------------------------------------------------------------------------------%
\section{mechanism analysis}\label{sec:mechanism}

\emph{Evolution of action, state, and reward.} 
To understand how the reversal in responder strategy unfolds, we fix the fairness reward $r=0.2$ and analyze three representative fairness weights ($\omega = 0.2$, $0.5$, $0.8$), which represent weak, intermediate, and strong fairness pressure, respectively. These three scenarios correspond to the regime at pre-reversal, within reversal, and post-reversal, respectively, and are illustrated in Fig.~\ref{fig:ts}. Specifically, we track the evolution of actions, states, and rewards for each scenario. A common observation in all scenarios is the decline of actions $p_h$ and $q_h$ after the transient. This is because the associated strategies typically result in frequent failures in division, yielding both low payoffs and poor fairness outcomes, and are thus ruled out in the evolution. 

%------------------------Fig. 5---------------------%
\begin{figure*}[htbp]
	\centering
	\includegraphics[width=0.9\textwidth]{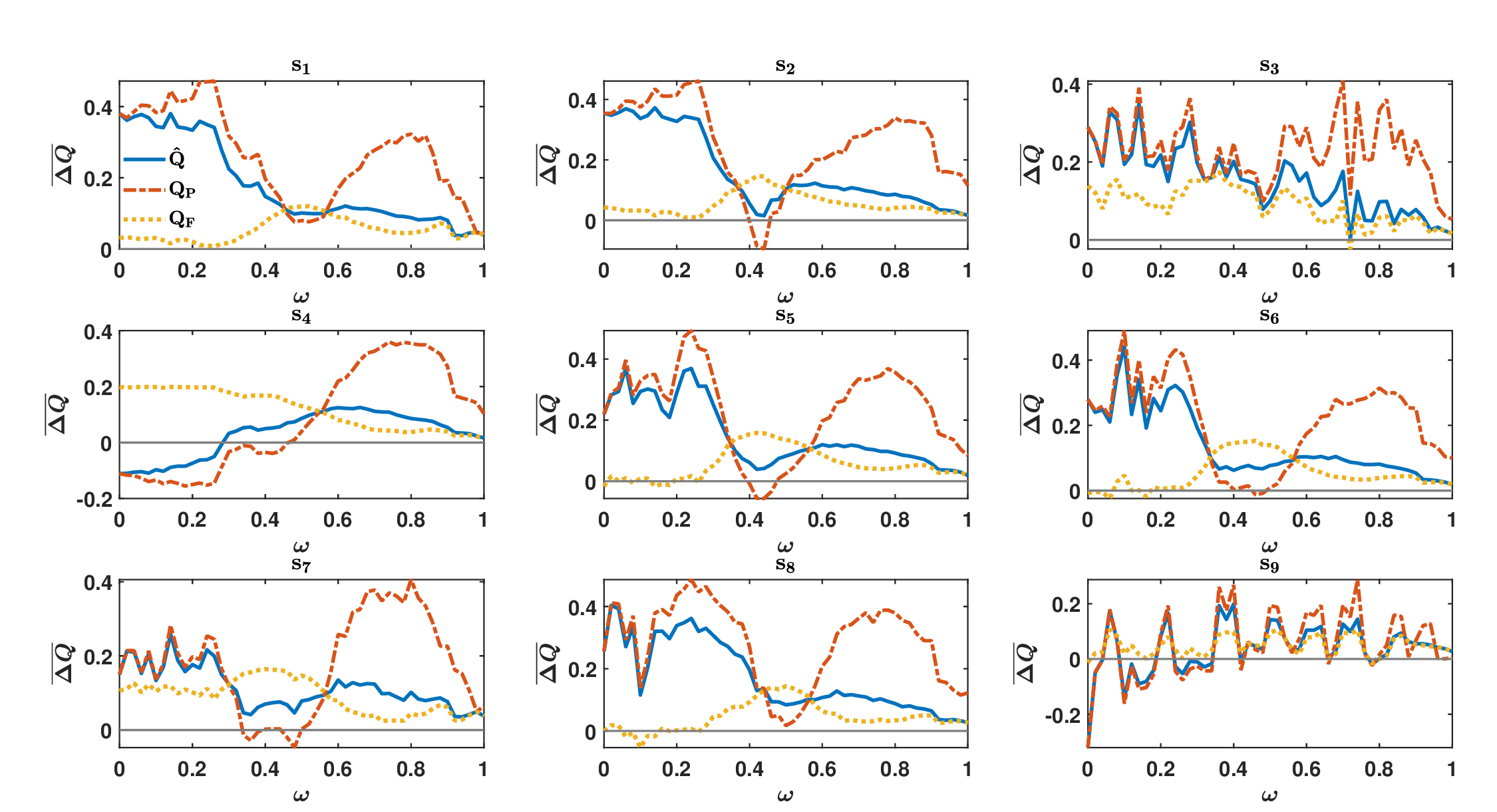}
	\caption{\textbf{Average Q-value advantage $\overline{\Delta Q_{s_i}}$ vs. $\omega$ across proposer states.}
		All data are obtained from an ensemble average over 100 independent runs. Other parameters: $\varepsilon = 0.01$, $\alpha = 0.1$, $\gamma = 0.9$, $t_{\mathrm{max}} = 1 \times 10^8$.}
	\label{fig:7}
\end{figure*}
%---------------------------------------------------%

At weak fairness pressure ($\omega = 0.2$), Fig.~\ref{fig:ts}(a) shows that the fair choices ($p_m$ and $q_m$) gradually dominate, but still a few low options ($p_l, q_l$) persist. The evolution of states shown in Fig.~\ref{fig:ts}(b) illustrates that there are basically only two combinations $(m,m)$ and $(l,l)$ left, meaning that most divisions are successful. But the split of $(l,l)$ lowers the responder's reward and yields unfairness, as confirmed in Fig.~\ref{fig:ts}(c). There, the average reward of responder $r_q$ is smaller than the proposer's $r_p$, and the average fairness reward $r_f$ is lower than the value $r=0.2$, expected in a fully fair scenario. These observations are qualitatively the same as the findings revealed in the previous single-objective scenario~\cite{Zheng2025DecodingFairness}, and weak fairness pressure fails to make a difference.

At intermediate fairness pressure ($\omega = 0.5$), Fig.~\ref{fig:ts}(d) shows that the proposer exclusively selects the fair offer $p_m$ driven by the fairness incentive, and the low offer $p_l$ vanishes. Interestingly, the responder chooses the low acceptance threshold $q_l$ more often than $q_m$. This can be interpreted as a ``forgiving" change towards an occasional low offer made by a fair proposer to guarantee a successful division. Fig.~\ref{fig:ts}(e) shows that the state combination $(m,l)$ is abundant, but the proposer prefers not to lower its offer to improve the payoff at the cost of reducing the fairness reward. Fig.~\ref{fig:ts}(f) shows that the average rewards for both proposer and responder are identical, and full fairness is now reached as $r_f=r$.

When the fairness pressure becomes strong ($\omega = 0.8$), being fair becomes the dominating driving force of evolution. The time series in Fig.~\ref{fig:ts}(h,i) show a similar evolution to Fig.~\ref{fig:ts}(d,e), but now the densities of $q_l$ and $q_m$ are reversed. This means the responder is now not that forgiving; she/he now cannot accept a low offer, even if this offer is occasional due to random exploration, e.g., by ``the trembling hands". As a result, the states mostly consist of $(m,m)$ with a few $(m,l)$. As expected, a fully fair outcome is reached, where the rewards of the two players become identical, shown in Fig.~\ref{fig:ts}(i).

%------------------------Fig. 6---------------------%
\begin{figure*}[htbp]
	\centering
	\includegraphics[width=0.9\textwidth]{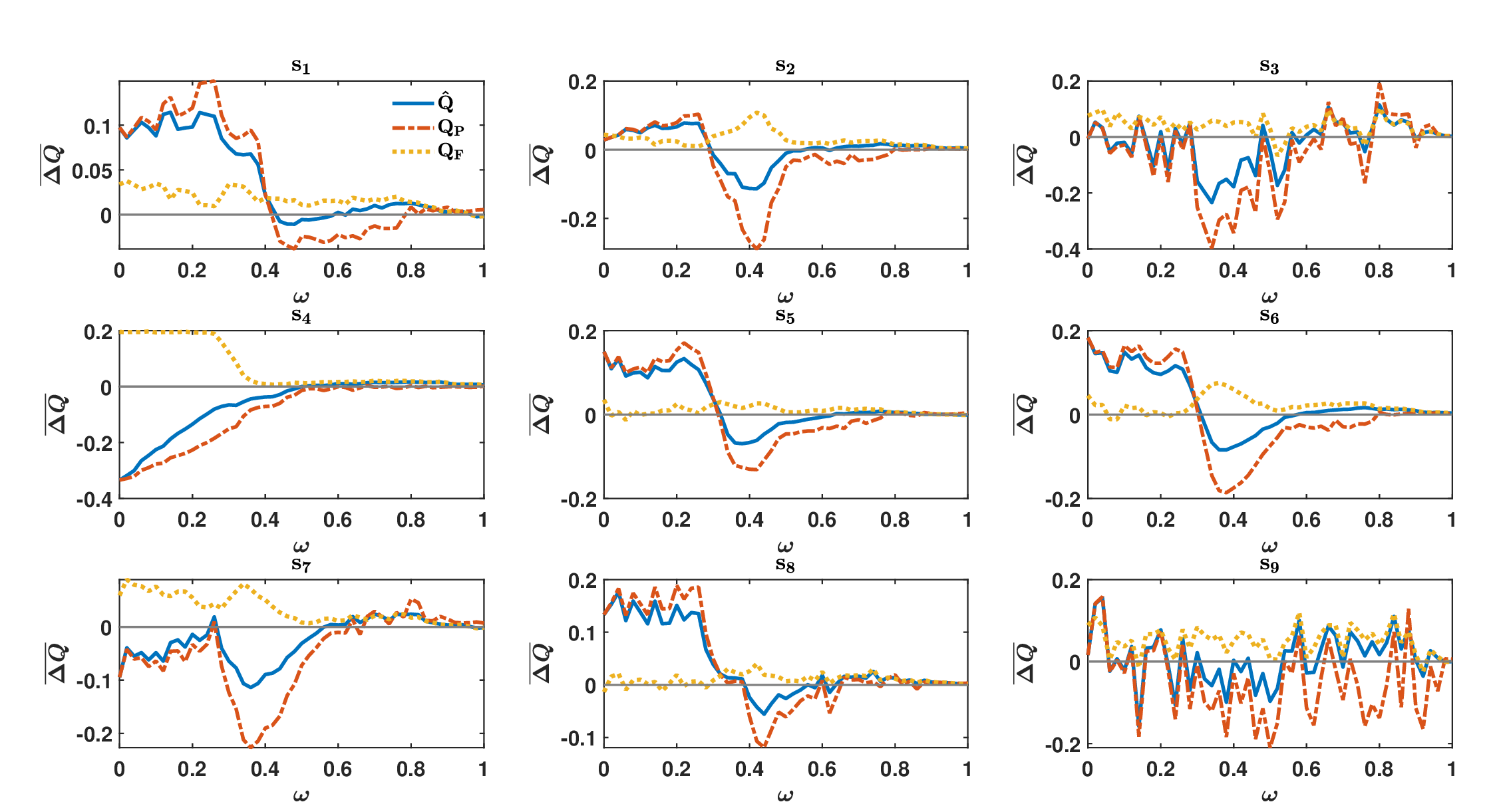}
	\caption{\textbf{Average Q-value advantage $\overline{\Delta Q_{s_i}}$ vs. $\omega$ across responder states.}
		All data are obtained from an ensemble average over 100 independent runs. Other parameters: $\varepsilon = 0.01$, $\alpha = 0.1$, $\gamma = 0.9$, $t_{\mathrm{max}} = 1 \times 10^8$.}
	\label{fig:8}
\end{figure*}
%----------------------------------------------------%

\emph{Analysis of Q-tables.} 
To uncover the microscopic learning mechanisms underlying the observations, we further examine the evolution of the Q-tables. Given that the high offer/acceptance threshold ($p_h$ and $q_h$) is virtually vanishing, we restrict our analysis to the low ($l$) and intermediate ($m$) actions, and distinguish the two roles. We define the following Q-value difference as the action preference of $m$ over $l$ within a given state $s_i$ as:
\begin{equation}\label{eq:r6}
	\overline{\Delta Q}_{s_i} = Q_{s_i, a_2(m)} - Q_{s_i, a_1(l)}.
\end{equation}
According to the working logic of Q-learning, $\overline{\Delta Q}_{s_i} > 0$ means that action $m$ is more preferred than action $l$ on average in state $s_i$; conversely, $\overline{\Delta Q}_{s_i} < 0$ indicates that action $l$ is preferred. 

Fig.~\ref{fig:7} provides the proposer's preference across the full spectrum of fairness pressure within all nine states, and we distinguish the preference in the three Q-tables, i.e., for the payoff $Q_P$, the fairness $Q_F$, and the aggregate Q-tables $\widehat{Q}$ defined by Eq.~(\ref{eq:r4}). We observe that the value of $\overline{\Delta Q}_{s_i}$ for the aggregate Q-tables is mostly positive, meaning $p_m$ is preferred. This explains the overall dominating prevalence of $p_m$ observed above [c.f. Fig.~\ref{fig:phase_diagram}(b)]. One exception is for the small $\omega$ within the state $s_4=(m,l)$, where the proposers are likely to provide a low offer $l$ when $\omega < 0.3$, as the responder's acceptance threshold is also low. But they turn to offer $m$ when the fairness pressure increases. In addition, there are some cases (e.g. $s_{2,5,7}$) where the proposer prefers $p_l$ to $p_m$ from the payoff Q-table $Q_R$, but the strong preference for $m$ from the fairness Q-table $Q_F$ ultimately forces the proposer to turn to the fair offer $p_m$.  %For exmaple, within the state of $s_2=(l,m)$ -- where the proposer previously submitted $l$ and the responder accepted with threshold $m$,  

The corresponding preference of the responder is shown in Fig.~\ref{fig:8}, which differs dramatically. At weak fairness pressure (small $\omega$), the responder typically prefers $q_m$, as most of the $\overline{\Delta Q}$ are positive, e.g., for the two dominating states $s_1=(l,l)$ and $s_5=(m,m)$ as shown in Fig.~\ref{fig:ts}(b). At intermediate fairness pressure, $\overline{\Delta Q}$ turns to be negative within the two dominating states $s_4=(m,l)$ and $s_5=(m,m)$ [Fig.~\ref{fig:ts}(e)]. At strong fairness pressure ($\omega\rightarrow 1$), we observe that $\overline{\Delta Q}\approx 0$; no clear preference is detected. This is understandable since the offer is $p_m$; a fair outcome is always reached for either of the three acceptance thresholds, even for $q_h$ where $\pi_{p}=\pi_r=0$. Put together, the preference changes within the three typical scenarios explain the reversal in $q$.
Notice that the preference in state $s_9=(h,h)$ is less obvious in the form of the fluctuating $\overline{\Delta Q}$; this is because this state is rarely visited and thus the preference is not well-developed.

%These findings demonstrate that agents' decision-making across the fairness-weight spectrum fundamentally reflects the dynamic interplay and competitive weighting of the payoff-derived and fairness-derived Q-components.

%----------------------------Extension----------------------------%
\section{extension}\label{sec:extension}

While the above two-objective setup applies to both proposer and responder symmetrically, a typical observation in behavioral experiments on the UG is often made that the two roles show distinct biases toward fairness.  
Specifically, experiments show that, compared to proposers, responders -- who are constrained to accept or reject -- are more sensitive to the division fairness~\cite{Cooper2011dynamics}. 
This means that the weight of the two objectives for the two roles could differ significantly. Moreover, there is asymmetry in the perceived information: responders can directly observe the proposer's offer, but proposers are not necessarily informed about the responder's acceptance threshold. 
To incorporate these two realistic features of human decisions, we revise the proposer's state representation from its previous joint action to the proposer's action in the last round, denoted as $\mathcal{S} = \mathcal{A} = \{ l, m, h \}$. The proposer's learning objective is also simplified to pure payoff maximization, i.e., the fairness incentive is no longer incorporated, $\omega=0$ in Eq. (\ref{eq:r4}). The responder, by contrast, keeps the two-objective Q-learning setup unchanged, and other parameters remain the same.

\begin{figure*}[htbp]
	\centering
	\includegraphics[width=1\linewidth]{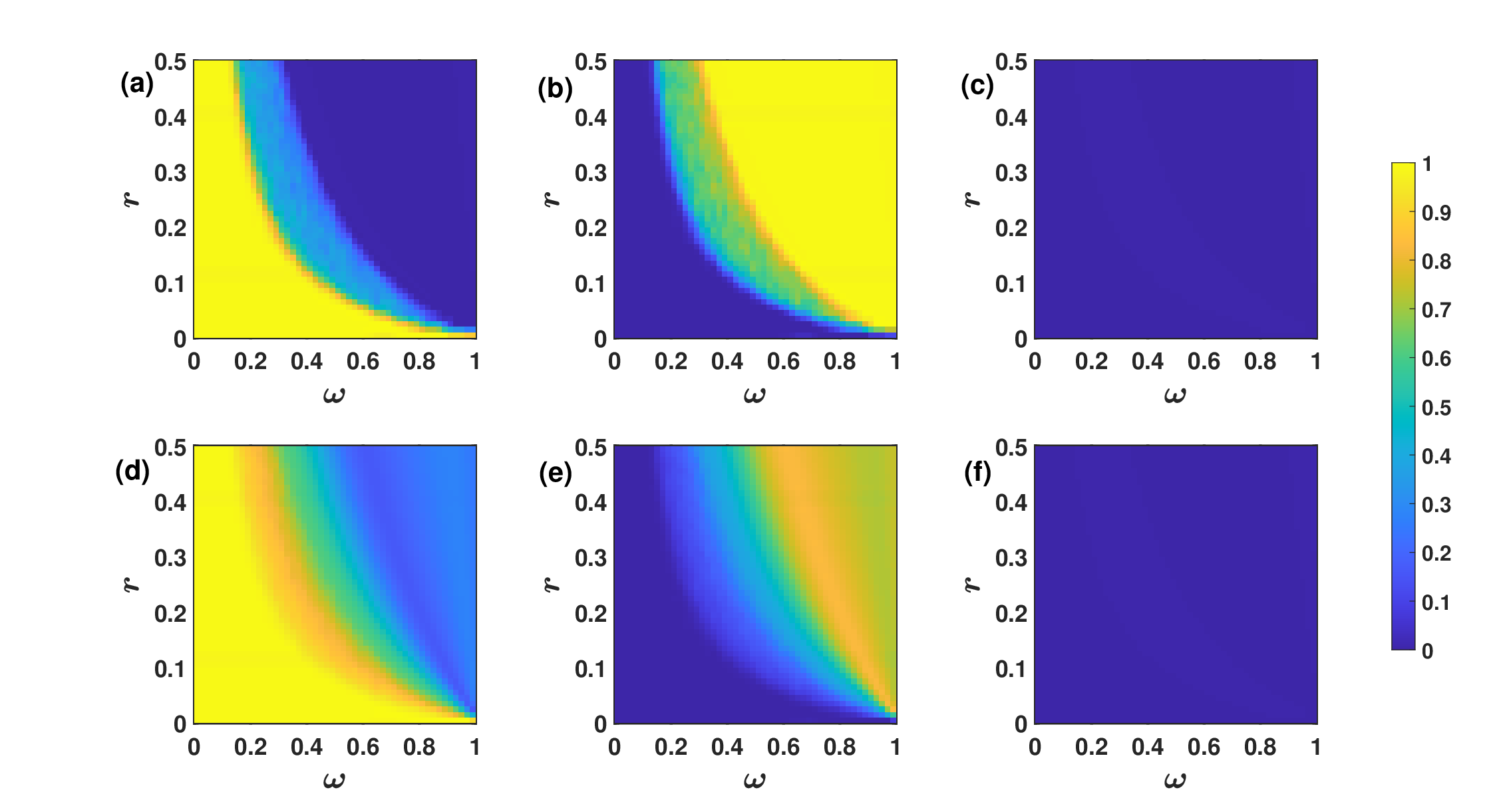}
	\caption{\textbf{Phase diagram of probabilities for the three actions with asymmetrical setup.} Here the proposer adopts the single-objective (i.e., payoff maximization) Q-learning, yet the responder is guided by the two-objective Q-learning as above. Probabilities of the action adoption for both proposer and responder are color-coded in the $r-\omega$ parameter space.
Panels (a-c) show the average proportion of the proposer's three actions for $p=l, m, h$, respectively. Panels (d-f) show the average proportion of the responder's three actions, respectively, for $q=l, m, h$. Each data point is obtained from an ensemble average over 100 independent runs. Other parameter: $t_{\text{max}} = 1 \times 10^8$.}
	\label{fig:asymmetry}
\end{figure*}

 Simulations with this asymmetric setup are presented in Fig.~\ref{fig:asymmetry}. At sufficiently low values of the fairness reward $r$ and the fairness weight $\omega$, both players converge to the low offer ($l$) as $\rho_{p_l,q_l}\rightarrow 1$ [Figs.~\ref{fig:asymmetry}(a,d)], yielding the Nash equilibrium predicted by the \emph{homo economicus} assumption. However, as the responder's fairness pressure (i.e., $r$ and $\omega$) increases, the adoption of fair options becomes more likely as $\rho_{p_m,q_m}\rightarrow 1$. Likewise, Figs.~\ref{fig:asymmetry}(c,f) show that the adoption probabilities $\rho_{p_h,q_h}\approx 0$ across the whole domain, which are different from Figs.~\ref{fig:phase_diagram}(c,f) where the fairness objective makes zero but equal payoffs at the extreme of $\omega=1$. The phase diagram shows that fair behavior can emerge solely from the responder's MORL, moving from completely unfair to fully fair as the two key parameters are changed. 
 
Importantly, the comparison of the single-objective case (i.e., $\omega=0$) with results in Fig.~\ref{fig:phase_diagram} and Ref.~\cite{Zheng2025DecodingFairness} shows that the asymmetric state setup here completely suppresses the emergence of fairness. This means that the perceived information as the state dramatically influences the game evolution, in line with previous studies~\cite{Zheng2025CooperationInformation}. A proper state setup requires further experimental evidence for accurate theoretical modeling.

%------------------------Fig. 7---------------------%
%---------------------------------------------------%

%------------------------------------------------------------------------------%
\section{conclusion}\label{sec:conclusion}

Motivated by the multidimensional nature of human decision-making, this work extends the prevailing single-objective reinforcement learning (RL) paradigm~\cite{Zheng2026Brief} to a multi-objective framework (MORL). In this framework, agents make decisions by navigating trade-offs among multiple, potentially conflicting objectives. Specifically, we apply the MORL framework to the ultimatum game to investigate the evolution of fairness with two objectives: material payoff maximization and fairness-driven moral incentives. Each objective is governed by a separate Q-table, and the two are integrated via a tunable fairness pressure coefficient into an aggregated Q-table that jointly guides the decision-making of both proposers and responders.

Our results show that the proposed two-objective learning framework promotes significantly higher fairness levels compared to the single-objective baseline that optimizes only material payoffs. Full fairness -- characterized by consistent equal splits -- emerges when the fairness pressure exceeds a moderate threshold. More interestingly, the responder exhibits non-monotonic strategic dynamics as fairness pressure increases, a pattern in line with our daily experience but lacking a formal computational account. Concretely, as pressure rises from low to moderate levels, the responder lowers the acceptance threshold to secure division success, even at the expense of fairness; further increases, however, restore a more stringent threshold, reflecting a renewed commitment to fairness. The promotion of fairness echoes previous studies~\cite{Zheng2022probabilistic,Zheng2023pinning}, which also show that occasional fair behaviors that deviate from the assumption of \emph{Homo economicus} can drive the whole population to a high fairness level.

As an emerging paradigm, RL offers a fresh lens for understanding a wide range of human social behaviors by emphasizing that many emergent properties can arise from endogenous incentives rather than exogenous factors. Nevertheless, this paradigm remains in its early stages and is not yet mature enough to serve as a comprehensive theoretical model for real-world human behavior.
By generalizing single-objective RL to a multi-objective formulation, our work provides a principled and extensible framework by capturing the intrinsic complexities of human sociality. We anticipate that this MORL approach will prove powerful and versatile in modeling not only the evolution of fairness, but also other foundational social phenomena such as cooperation, trust, and more.

%------------------------------------------------------------------------------%
\section*{Data and code availability}
The code for generating key results in this study is available at \href{https://github.com/chenli-lab/MORL}{https://github.com/chenli-lab/MORL}.

\section*{ACKNOWLEDGEMENTS}
This work is supported by the National Natural Science Foundation of China (Grants Nos. 12675043, 12165014), the Fundamental Research Funds for the Central Universities (Grant Nos. GK202401002 and GK202406016), and the Natural Science Basic Research Program of Shaanxi (Grants Nos. 2026JC-YBMS-0012, 2026JC-YBQN-0021).

\bibliography{reference}

\end{document}